\documentclass[10pt,twocolumn,letterpaper]{article}

\usepackage{iccv}
\usepackage{times}
\usepackage{epsfig}
\usepackage{graphicx}
\usepackage{amsmath}
\usepackage{amssymb}
\usepackage[table, usenames, dvipsnames]{xcolor}
\usepackage{booktabs} 
\usepackage{arydshln}
\usepackage{multirow}
\usepackage{xcolor}
\usepackage{lipsum}
\usepackage{comment}
\usepackage{enumitem}
\usepackage{subcaption}

\usepackage{enumerate}
\usepackage{algpseudocode}
\usepackage{algorithm}

\usepackage[pagebackref=true,breaklinks=true,letterpaper=true,colorlinks,bookmarks=false]{hyperref}

\iccvfinalcopy

\ificcvfinal\fi

\begin{document}

\title{Unsupervised Robust Domain Adaptation without Source Data}

\author{Peshal~Agarwal$^{1}$\quad Danda~Pani~Paudel$^{1}$\quad Jan-Nico~Zaech$^{1}$\quad Luc~Van~Gool$^{1,2}$\\
$^{1}$Computer Vision Laboratory, ETH Zurich, Switzerland \quad $^{2}$KU Leuven, Belgium\\
\tt\small{agarwalp@student.ethz.ch}\quad{\tt\small\{paudel, zaechj, vangool\}@vision.ee.ethz.ch}
}

\maketitle
\ificcvfinal\thispagestyle{empty}\fi

\begin{abstract} 

We study the problem of robust domain adaptation in the context of unavailable target labels and source data. The considered robustness is against adversarial perturbations. This paper aims at answering the question of finding the right strategy to make the target model robust and accurate in the setting of unsupervised domain adaptation without source data. The major findings of this paper are: (i) robust source models can be transferred robustly to the target; (ii) robust domain adaptation can greatly benefit from non-robust pseudo-labels and the pair-wise contrastive loss. The proposed method of using non-robust pseudo-labels performs surprisingly well on both clean and adversarial samples, for the task of image classification.
We show a consistent performance improvement of over $10\%$ in accuracy against the tested baselines on four benchmark datasets. Our source code will be made publicly available. 
\end{abstract}
\section{Introduction}

Transferring the knowledge learned in one domain to another, in an unsupervised manner, is highly desired for a wide range of applications for learning-based methods~\cite{pan2010domain,torralba2011unbiased,ganin2015unsupervised,tzeng2017adversarial,wilson2020survey}. Many of these applications also require models to be robust towards data perturbations~\cite{papernot2016limitations,papernot2016distillation,kurakin2016adversarial}. In practice the source data may no longer be accessible during the knowledge transfer, due to privacy, storage or communication overhead;
An example where such limitations are clearly manifested is image understanding with datasets\footnote{Our work focuses on the image classification problem.} containing people's faces.
These practical limitations call for methods that can adapt using only the target data.  
Furthermore, adaptation without the need of source data 
can also offer benefits in terms of  
computational cost and may simplify data handling. 

This paper addresses a real world compound problem of (i) unsupervised domain adaptation; (ii) model robustness; and (iii) the lack of source data during transfer. All three mentioned issues are jointly considered, leading to a realistic yet a very challenging problem. 
Up to our knowledge, this problem is addressed for the first time in this work. 
\begin{figure}
    \centering
    \includegraphics[width=0.44\textwidth]{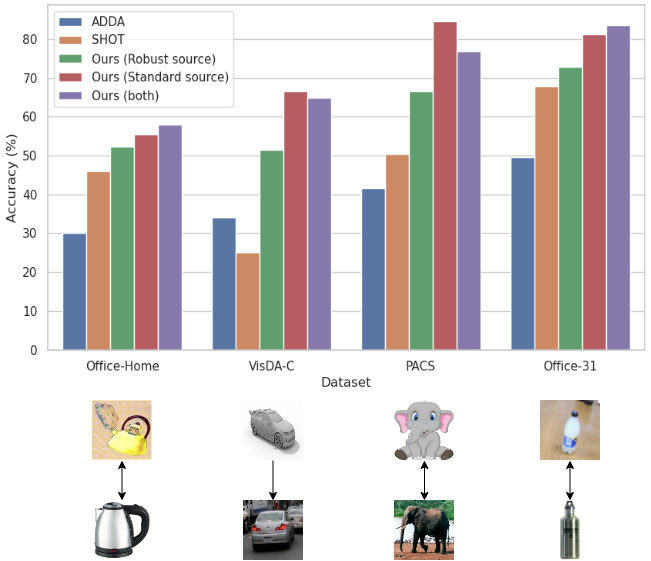}
    \caption{Test accuracy averaged over all domain adaptation tasks for multiple datasets. All our proposed methods show significant improvement over the baselines.}
    \label{fig:results}
\end{figure}

We consider that the source data is available only during the source model training, which is performed in a supervised manner. The model is then adapted to the target domain in an unsupervised manner, when the source data is no longer available. In this process, the robustness of the target model towards the adversarial perturbations is pursued.

This paper studies several aspects of designing a robust domain adaptation method and proposes a simple yet novel technique to answer the key questions of:

\begin{itemize}
    \setlength{\itemsep}{0pt}
    \setlength{\parskip}{0pt}
    \item How to perform robust and unsupervised domain adaptation without source data?
    \item Can robust and standard models be combined to efficiently use information from the source domain?
    \item How do we perform robust domain adaptation, if only one model (robust or standard) is available?
    \item Is the best adaptation approach dataset dependent?
\end{itemize}

The problem of unsupervised domain adaptation without source data has recently been studied in ~\cite{liang2020we,kim2020domain,sahoo2020unsupervised,li2020model,kurmi2021domain,kundu2020universal,yang2020unsupervised}. However, the existing work does not take robustness into consideration. In this work, we first show that robust domain adaptation performs reasonably well within the aforementioned setup, when the method of~\cite{liang2020we} is directly applied. The method exploits the target's pseudo-labels, generated by the source model, for adaptation, which  inevitably leads us to use robust pseudo-labels. We first study the performance of~\cite{liang2020we} under the adversarial perturbations for robustness, and then improve the performance by over $20\%$ in accuracy consistently across four benchmark datasets, as shown in Fig.~\ref{fig:results}. Such improvement is achieved by exploiting non-robust pseudo-labels and the target's pair-wise contrastive learning scheme. The major finding of our work is that \emph{robust domain adaptation can largely benefit from non-robust pseudo-labels and the pair-wise contrastive loss.} Our finding allows us to improve not only the robust accuracy, but also the clean accuracy in the target domains of a single robust model.

\begin{figure}
    \centering
    \includegraphics[scale=0.6]{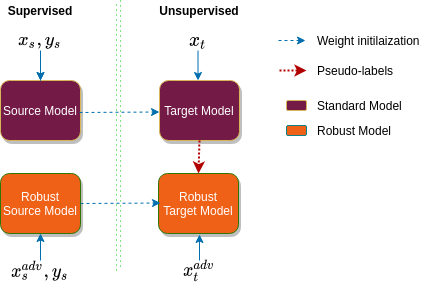}
    \caption{
    The training in the source domain uses the source data $x_s$, labels $y_s$, and adversarial examples $x_s^{adv}$. The training in the target domain uses the target data $x_t$ and the adversarial examples $x_t^{adv}$ generated using the pseudo labels.  
    }
    \label{fig:model-abs-se}
\end{figure}

We study three different cases of model availability: (i) given only the standard source model; (ii) given only the robust source model; (iii) given both models. In the following, we will first present the case when both models are available. In this case, we wish to adapt the robust source model to the target while guarding the robustness.    
During the adaptation of the robust model, the labels generated by the standard one are used in three different ways: (i) cross-entropy loss; (ii) adversarial examples generation; (iii) contrastive  feature learning. 
These three aspects of utility have shown to be complimentary to each other. Exploitation of non-robust pseudo-labels in this fashion also offer a significantly better performance compared to its robust counterpart. In fact, this observation leads us to suggest a new source data training and model sharing protocol. In the source domain, we suggest to train two models; one being robust and the other not. As shown in Fig.~\ref{fig:model-abs-se}, the transfer process utilizes both models, while the source data is not required during transfer. However, once the model has been adapted, only the robust model is required for inference.

Our main contributions are threefold:
\begin{itemize}
    \setlength{\itemsep}{0pt}
    \setlength{\parskip}{0pt}
    \item We study a new problem of unsupervised robust domain adaptation in the setting of missing source data.
    \item A simple yet a very effective method is proposed, which exploits the non-robust pseudo-labels for robustness, to address the problem at hand. 
    \item The proposed method is extensively tested on four benchmark datasets, consistently demonstrating the excellent improvements of over $10\%$ in accuracy.  
\end{itemize}

\section{Related Works}
\noindent\textbf{Unsupervised Domain Adaptation (UDA).}
Unsupervised domain adaptation is a topic of broad interest~\cite{saenko2010adapting,pan2010domain,torralba2011unbiased, gong2012geodesic,venkateswara2017deep,peng2019moment}. UDA aims at transferring supervised source domain models to an unlabeled target domain. The traditional UDA works \cite{long2015learning, ganin2015unsupervised, long2017deep, tzeng2017adversarial} typically focus 
solving adaptation problem using the source data, while being oblivious to the adversarial attacks. Despite of being very insightful, the traditional UDA methods are restricted in many practical settings due to the considered assumptions.

\noindent\textbf{UDA without Source Data.} The UDA without source data can be broadly divided into three categories: (i) generative approach~\cite{li2020model,kurmi2021domain,kundu2020universal}; (ii) pseudo-label approach~\cite{kim2020domain,liang2020we}; and (iii) others~\cite{sahoo2020unsupervised,yang2020unsupervised}.
The generative approach is often difficult to scale up, as learning to generate the images/features is known to be difficult. On the other hand, pseudo-label based method are easy to handle and have recently provided very promising results~\cite{liang2020we}. The third category of the methods are either designed under very simplistic settings (\cite{sahoo2020unsupervised} is dedicated to the pixel level corruptions) or demand sophisticated mechanism without offering significant gain over the pseudo-label based methods. Therefore, we  also leverage the pseudo-labels as in~\cite{kim2020domain,liang2020we}.

\begin{figure*}[t!]
    \centering
    \includegraphics[scale=0.65]{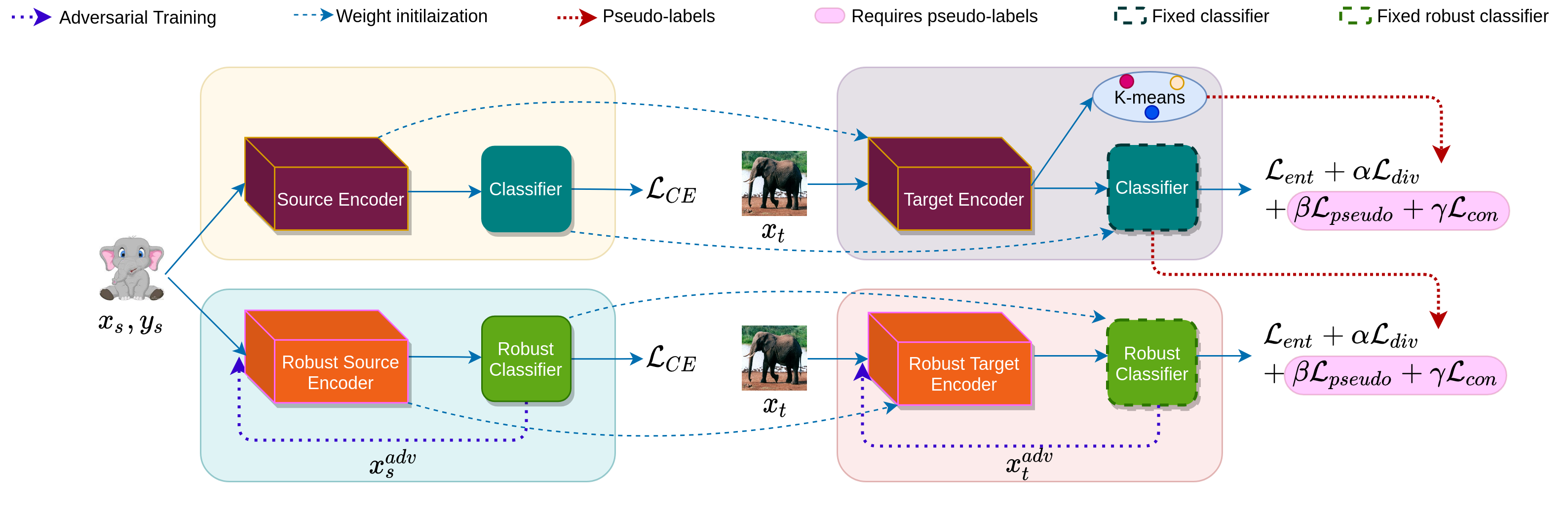}
    \caption{First, a standard (top-left) and a robust model (bottom-left) are trained on source. Then, a target encoder (top-right) is trained by combining four losses with pseudo-labels that are obtained via k-means. Finally, a robust target encoder (bottom-right) is trained similarly to standard target with two modifications. One, the pseudo-labels are obtained from the pre-trained standard target model. Two, adversarial images are generated to facilitate adversarial training.}
    \label{fig:model_sep}
\end{figure*}

\noindent\textbf{Robust Training.}
A flurry of attack mechanisms~\cite{carlini2017towards, chen2017zoo, brendel2017decision, nayebi2017biologically, papernot2016distillation, baluja2017adversarial, luo2018towards} has been proposed since the vulnerability was shown first  by \emph{Goodfellow et al.}~\cite{goodfellow2014explaining}. This has also lead strategies that can defend against such attacks, called defense mechanisms~\cite{jang2019adversarial, liao2018defense, papernot2016distillation, samangouei2018defense, akhtar2018defense, schott2018towards, mustafa2019adversarial}. Among them, adversarial training~\cite{goodfellow2014explaining, kurakin2016adversarial} has stood out as the most reliable way to train robust models.
We follow the adversarial training method proposed by \emph{Madry et al.}~\cite{madry2017towards} because of being effective, fast, and easy to implement.

\noindent\textbf{Robust Transfer.}
Our work is also inspired from the recent works on robust transfer learning~\cite{shafahi2019adversarially, salman2020adversarially, utrera2020adversarially} in supervised settings. A notable work of \emph{Shafahi et al.}~\cite{shafahi2019adversarially} shows that a robust source feature extractor can be effective in preserving robustness, while maintaining \textit{sufficiently high} accuracy on the clean samples. On the other hand, \cite{salman2020adversarially, utrera2020adversarially} show that the robust pre-trained models not only perform well on targets without adversarial training, but also improve the accuracy on clean samples. These results strengthen the hypothesis that robust models also transfer better. However, the existing methods are neither developed or tested in the settings of unsupervised domain adaptation.

\section{Robust Adaptation} \label{sec:method}
In the following we elaborate our methodology for unsupervised robust domain adaptation for multi-class classification problem without access to the source data. Given a dataset \{$(x^1_s, y^1_s), (x^2_s, y^2_s), \dots, (x^n_s, y^n_s)\}$ where $(x^i_s, y^i_s) \sim \mathcal{D}_s$ comes from the source domain, our goal is to train a model that can predict target labels $y_t$ for the corresponding target images $x_t$ where $(x_t, y_t) \sim \mathcal{D}_t $ and is robust to adversarial examples at the same time. We can broadly separate the process into two phases. In the initial phase, we train a model on the source domain in a supervised fashion, and in the final phase, we adapt the model to the target domain. Formally, we need to learn a function $f_s: X_s \rightarrow Y_s$ on the source domain and use that information along with the target data to learn another function $f_t: X_t \rightarrow Y_t$. To this end, we train two models in each domain (source and target), one of them following the standard protocol and one being robust to adversarial examples. We propose to train four models in total: source model, robust source model, target model, and robust target model, as shown in Figure \ref{fig:model-abs-se}.
For the sake of brevity, we will refer to pseudo-labels obtained from {standard} and {robust models} as \emph{non-robust pseudo-labels} and \emph{robust pseudo-labels}, respectively.

\subsection{Source Training}
A deep neural network is trained on the source domain by minimizing the standard cross-entropy loss given by,
\begin{equation}
\mathcal{L}_s(f_s; X_s, Y_s) = \mathbb{E}_{(x_s, y_s) \sim \mathcal{D}_s} \mathcal{L}_{CE}(f_s(x_s), y_s). 
\end{equation}
Besides a standard source model, we also train a robust source model. Here, the objective is to learn a function on the source domain that is robust to adversarial images. We generate adversarial perturbations $\eta$ under the $l_{\infty}$ threat model. This leads to minimizing the worst case cross-entropy loss, within the $l_{\infty}$ ball of fixed radius, as follows,
\begin{equation}
\small
\mathcal{L}^r_s(f_s; X_s, Y_s) = \mathbb{E}_{(x_s, y_s) \sim \mathcal{D}_s} \underset{x' \in S(x_s)}{\max} \mathcal{L}_{CE}(f_s(x'), y_s),
\end{equation}
where $S(x) = \{x'\ |\ ||x - x'||_\infty < \epsilon\}$ and $\epsilon$ is the perturbation threshold.
Note that finding a sample within $S(x)$ that maximizes the cross-entropy is computationally challenging due to infinitely many samples in $S(x)$ and no closed-form solution. Thus, we empirically generate adversarial examples using Projected Gradient Descent (PGD) and perform adversarial training~\cite{madry2017towards}.

Both standard and robust source models described in Figure \ref{fig:model_sep} have two components, namely, an encoder and a classifier. We will use $\Phi_s : X_s \rightarrow \mathbb{R}^d$ and $\Phi^r_s : X_s \rightarrow \mathbb{R}^d$ to denote the standard and robust source encoders, respectively. Similarly, the corresponding classifiers are denoted as, $\delta: \mathbb{R}^d \rightarrow \mathbb{R}^C$ and $\delta^r : \mathbb{R}^d \rightarrow \mathbb{R}^C$, for feature dimension $d$  and $C$ classes.
We will make use of only two classifiers for both source and target domains. 
Both classifiers are trained on the source data, and will remain unchanged for the target, similar to~\cite{liang2020we}.

\subsection{Target Training} 
Our target training stage assumes that only the source trained models are available. Furthermore, the target data are provided without class labels.
Our method for target only training is inspired from the source hypothesis transfer~\cite{liang2020we}, which has shown impressive performance on the standard unsupervised domain adaptation. In this work, we extend \cite{liang2020we} to the case of robust model adaptation. 
Similar to the source domain, our approach relies on two separate models in the target domain, namely, the standard and robust target models. We initialize the weights of each model with the corresponding source models. During the adaptation process, the encoders are optimized while keeping the classifier fixed. In the target domain, standard and robust models are trained differently. In the following, we will first present our approach for training standard model followed by the same for the robust model. The key aspects of our method is summarized in Algorithm~\ref{alg:ourAlgorithm}.
\subsubsection{Standard Model}
\label{subSec:nonrobustTraning}
The idea of standard training is to learn a standard target encoder $\Phi_t : X_t \rightarrow \mathbb{R}^d$ that generates features which align closely with the corresponding source feature distribution, making it possible to re-use the source classifier $\delta(.)$. No access to source data restricts us to perform direct alignment between the two features as in~\cite{tzeng2017adversarial}.
To address this problem, our approach involves (i) entropy and divergence of the predicted labels, (ii)  pseudo-label based supervision, and (iii) contrastive target features. The first two aspects are borrowed from ~\cite{liang2020we} and other prior works~\cite{wang2020fully,shannon1948mathematical}. The aspect of using contrastive feature learning is proposed in this work, for the first time to address the problem at hand. 

\noindent\textbf{Entropy and Divergence:} Entropy minimization is a widely used technique for unsupervised domain adaptation~\cite{wang2020fully}. The Shannon entropy~\cite{shannon1948mathematical} for a prediction probability $\hat p_i$ of class $i$ is defined as, 
\begin{equation}\label{eq:entropy}
    \mathcal{L}_{ent} = - \underset{i}{\Sigma}\ \hat p_i \log \hat p_i.
\end{equation}
Unfortunately, entropy minimization can produce degenerate labels with loss converging to zero. Therefore, we take the information maximization (IM)~\cite{gomes2010discriminative} approach as adopted by~\cite{liang2020we}. 
IM adds an additional diversity term that pushes the predicted labels to be uniformly distributed avoiding the trivial outcome of the same one-hot vector for all inputs. Let $q_i$ be the average probability of a prediction for the class $i$, then the diversity loss is defined as, 
\begin{equation}\label{eq:diversity}
    \mathcal{L}_{div} =  \underset{i}{\Sigma}\ q_i \log q_i.
\end{equation}

\noindent\textbf{Non-robust Pseudo-labels:}
While IM can make the model confident while ensuring diverse prediction, it may still push the output towards incorrect prediction in certain cases. In order to overcome such undesired behaviour,~\cite{liang2020we} proposed to use pseudo-labels~\cite{lee2013pseudo} in addition to IM for better supervision. We use two-step weighted k-means clustering on the feature space to obtain pseudo-labels as described in~\cite{liang2020we}. Let $\hat y$ be the pseudo-label obtained for the image $x$. Then, the pseudo loss is defined using the cross-entropy as,
\begin{equation}\label{eq:pseudo}
    \mathcal{L}_{pseudo} = \mathcal{L}_{CE}(\delta(\Phi_t(x)), \hat y).
\end{equation}

\noindent\textbf{Constrastive Feature Learning:} We use the obtained  pseudo-labels also to learn the discriminative features in the target. The proposed use of the contrastive loss is inspired by~\cite{sohn2016improved, khosla2020supervised}, which were originally used in different contexts. The contrastive loss minimizes the intra-class distance, while maximizing inter-class distance between the encoder features. For two input images $x_1, x_2$ with pseudo-labels $y_1, y_2$, the pair-wise contrastive loss is given by,
\begin{equation}\label{eq:contra}
    \mathcal{L}_{con} = \frac{1}{2}[y\cdot D^2 + (1 - y) \cdot \max(0, m - D)^2],
\end{equation}
where $y = \mathbb{I}_{\{y_1 = y_2\}}$,  $D = ||\Phi_t(x_1) - \Phi_t(x_2)||_2$ and $m > 0$ is the margin between features of different classes.

To optimize the target standard model, we minimize the weighted combination of the loss terms described above. In this context, the minimized loss is given by,
\begin{equation}\label{eq:target}
    \mathcal{L}_t(f_t; X_t, Y_t) = \mathcal{L}_{ent} + \alpha \mathcal{L}_{div}
                                    + \beta \mathcal{L}_{pseudo} + \gamma \mathcal{L}_{con},
\end{equation}
where $\alpha,\beta$, and $\gamma$ are the weights corresponding to the respective loss functions. 

\subsubsection{Robust Model}\label{subSec:robustTraning}
The idea of robustness transfer is inspired by some of the recent works~\cite{shafahi2019adversarially, utrera2020adversarially, salman2020adversarially} in this direction.
Some existing works perform the knowledge transfer using a robust source model.
Such transfer is shown to preserve the robustness also for the new tasks.
In our work, we show that the robust source model also transfers robustly to the target, up to some extent. To improve the robustness further, we propose adversarial training also in the target. Unfortunately, the adversarial robust training often requires labeled examples. 
One may consider using the pseudo-labels from the robust model.
However, due to the trade-off between clean and robust accuracy~\cite{zhang2019theoretically}, this process will result into less accurate pseudo-labels. Instead, we propose to obtain the required pseudo-labels using the standard model. Note that the clean accuracy of the standard models is higher than that of the robust ones. More importantly, the pseudo-labels obtained using a standard model, for clean samples, are sufficient to generate the required adversarial examples. 

At this point, we wish to transfer the source robustness using a robust source model. On the other hand, we require better pseudo-labels to generate adversarial examples. Therefore, we use both robust and standard source models and transfer them to the target domain. In this process, the robustness of the robust model is reinforced by using the pseudo-labels from the standard model. 
Additionally, we believe that the used pseudo-labels offer better domain alignment by means of minimizing the cross-entropy and pair-wise contrastive losses of ~\eqref{eq:pseudo} and ~\eqref{eq:contra}, respectively.

\noindent\textbf{Adversial Target Examples:}
We generate adversarial examples using PGD method~\cite{madry2017towards}. 
These generated images are used to compute the IM loss of~\eqref{eq:entropy} and~\eqref{eq:diversity}. 
We train two models independently on the target domain.
The standard model is trained first, followed by the robust one. 
The final loss use for the robust training is given by,
\begin{equation}\label{eq:target_robust}
    \mathcal{L}^r_t(f^r_t; X_t, Y_t) = \mathcal{L}^r_{ent} + \alpha \mathcal{L}^r_{div}
                                    + \beta \mathcal{L}^r_{pseudo} + \gamma \mathcal{L}^r_{con}.
\end{equation}

 \begin{algorithm}
 \caption{Target adaption using two models. \label{alg:ourAlgorithm}}\vspace{0.5mm}
    \begin{algorithmic}[1]
    \State Initialize weights of $\Phi_t(.)$ with $\Phi_s(.)$
    \For{$epoch < MaxEpochs$}
        \State Obtain pseudo-labels $\hat y$ via k-means
        \For{each mini-batch}
            \State Update the $\Phi_t(.)$  using Eq~\eqref{eq:target}
        \EndFor
        \If{$epoch\ \%\ update = 0$}
        \State Update the pseudo-labels
        \EndIf
    \EndFor
    \State Initialize the weights $\Phi^r_t(.)$ with $\Phi^r_s(.)$
    \State Obtain pseudo-labels $\hat y$ via $\delta(\Phi_t(x))$
    \For{$epoch < MaxEpochs$}
        \For{each mini-batch }
            \State Obtain $x^{adv}_t$ for $x_t$ using $\hat y$ and $\delta(\Phi^r_t(x))$
            \State Update the $\Phi^r_t(.)$  using Eq~\eqref{eq:target_robust}
        \EndFor
    \EndFor
    \end{algorithmic}
\end{algorithm}

\subsection{Adaptation with a Single Source Model}\label{subSec:singleModel}
The method previously presented suggests a model hand over protocol, where the user with the access to the source data provides two models.
In some practical scenarios however, both models may not be available. Under such circumstances, we suggest to still \emph{use pseudo-labels with the proposed method for the best outcome of robust adaptation}, irrespective of the model being robust or standard. This suggestion is supported by our extensive experiments.  We will present these results as, (i) \textbf{Robust source}: uses only the robust source model,
(ii) \textbf{Standard source}: uses only the standard source model, 
(iii) \textbf{Both}: uses both models. In the source robust case, the robust pseudo-labels are used for adaptation. In the other two cases, non-robust pseudo-labels are used. Needless to say, the source standard case adapts the standard model robustly to the target domain.

\begin{table*}
    \begin{tabular}{c|c c c c c c c c}
    \hline
    \centering
    \multirow{2}{*}{Method} &
    \multicolumn{2}{c}{\textbf{Office-31}~\cite{saenko2010adapting}} &
    \multicolumn{2}{c}{\textbf{Office-home}~\cite{venkateswara2017deep}} &
    \multicolumn{2}{c}{\textbf{PACS}~\cite{li2017deeper}} &
    \multicolumn{2}{c}{\textbf{VisDA-C}~\cite{peng2017visda}} \\
    & {Adv acc} & {Clean acc} & {Adv acc} & {Clean acc} & {Adv acc} & {Clean acc} & {Adv acc} & {Clean acc}\\
    \hline \\[-1em]
    ADDA~\cite{tzeng2017adversarial}&  0.4 & 75.0 &  0.9 & 50.2 &  0.8 & 64.6 &  0.9 & 70.1 \\
    \\[-1em]
    SHOT~\cite{liang2020we} &  0.0 & \textbf{87.6} &  0.3 & \textbf{65.8} &  0.0 & 57.0 &  0.3 & \textbf{79.0} \\
    \\[-1em]
    ADDA \textit{robust}    & 49.6 & 57.6 & 30.1 & 37.7 & 41.6 & 59.8 & 34.0 & 46.3 \\
    \\[-1em]
    SHOT \textit{robust}    & 67.6 & 73.1 & 46.1 & 53.1 & 50.5 & 57.3 & 25.0 & 34.3 \\
    \\[-1em]
    \hline \\[-1em]
    Ours (Robust source)    & 72.8 & 76.4 & 52.4 & 59.2 & 66.5 & 72.7 & 51.4 & 63.6 \\
    \\[-1em]
    Ours (Standard source)  & \underline{81.2} & 85.7 & \underline{55.4} & 62.7 & \textbf{84.6} & \textbf{89.4} & \textbf{66.7} & \underline{75.8} \\
    \\[-1em]
    Ours (Both)             & \textbf{83.5} & \underline{87.0} & \textbf{58.0} & \underline{65.1} & \underline{76.9} & \underline{83.6} & \underline{65.0} & 74.9 \\
    \hline
    \end{tabular}
    \caption{Accuracy on adversarial and clean images on the test data averaged over all domain adaptation. All our methods have higher adversarial accuracy compared to the baselines. The performance of our methods on clean samples is comparable and mostly higher than the other methods. The best accuracy is presented in bold and the second best is underlined.}
    \label{table:results}
\end{table*}

\begin{table*}
    \centering
    \resizebox{\textwidth}{!}{%
    \begin{tabular}{c c c c c c c c c c c c c c}
         \hline
         Method & A $\rightarrow$ C & A $\rightarrow$ P & A $\rightarrow$ S & C $\rightarrow$ A & C $\rightarrow$ P & C $\rightarrow$ S & P $\rightarrow$ A & P $\rightarrow$ C & P $\rightarrow$ S & S $\rightarrow$ A & S $\rightarrow$ C & S $\rightarrow$ P & \textbf{Avg.} \\
         \hline
         w/o Contrastive & 73.8 & 93.4 & 41.1 & 68.0 & 83.8 & 42.9 & 71.0 & 60.1 & 29.4 & 6.6 & 22.0 & 17.1 & 50.8 \\
         w/o Cross-entropy & 92.3 & 93.4 & 48.3 & 78.5 & 91.9 & 65.4 & 82.4 & 50.3 & 37.2 & 2.4 & 8.7 & 2.7 & 54.5 \\
         w/o Entropy & 88.7 & 94.3 & 43.5 & 77.8 & 88.0 & 52.5 & 82.2 & 67.4 & 40.6 & 14.4 & 55.0 & 41.6 & 62.2 \\
         w/o Adv. Images & 87.0 & 93.7 & 24.9 & 76.3 & 92.8 & 21.5 & 74.6 & 88.3 & 17.3 & \textbf{76.1} & \textbf{91.5} & \textbf{54.2} & 66.5 \\
         w/o Diversity loss & \textbf{96.2} & \textbf{99.7} & 90.2 & \textbf{88.8} & \textbf{99.1} & 89.6 & \textbf{93.9} & 81.0 & 33.0 & 22.2 & 59.1 & 30.8 & 73.6 \\
         \textbf{Ours} (both) & 92.1 & 92.8 & \textbf{94.9} & 77.3 & 91.6 & \textbf{95.2} & 78.8 & \textbf{94.2} & \textbf{71.0} & 30.7 & 84.2 & 20.4 & \textbf{76.9} \\
         \hline
    \end{tabular}}
    \caption[Ablation Study]{Ablation study of our (both) target model on PACS datatset. The contrastive loss term, entropy term and diversity loss term were removed from both (standard and robust) the target models while Cross-entropy term was only removed from the target robust model since removing from both will make it very hard to adapt.}
    \label{table:ablation}
\end{table*}
\section{Experiments}\label{sec:experiment}
\subsection{Experimental Setup}
    \noindent\textbf{Datasets.}
    We conduct experiments on four benchmark datasets, including one small, two medium  and one large-scale dataset. The datasets vary in their number of classes from 7 to 65 and contain between two and four different domains. Office-31~\cite{saenko2010adapting} consists of total 4,652 images from three domains -~Amazon~(\textbf{A}), DLSR~(\textbf{D}) and Webcam~(\textbf{W})~- each having 31 classes. Office-home~\cite{venkateswara2017deep} is collected in four different domains -~Art~(\textbf{Ar}), Clipart~(\textbf{Cl}), Product~(\textbf{Pr}) and Real-world~(\textbf{Rw})~- each with 65 classes a total of 15588 images in the dataset. PACS~\cite{li2017deeper} contain 9991 images from four domains -~Art~(\textbf{A}), Clipart~(\textbf{C}), Photo (\textbf{P}) and Sketch (\textbf{S})~- where each image belongs to one of 7 classes. The largest considered dataset, VisDA-C~\cite{peng2017visda} has only two domains -~Synthetic~(\textbf{S}) and Real~(\textbf{R})~- with ~152k and ~55k images respectively. Therefore, each of the 12 different classes has a significantly larger number of samples than in the other datasets. For all datasets and all the adaptation tasks, we randomly split both the source and the target domain samples into train/val/test (0.7/0.1/0.2).
    
    \noindent\textbf{Network Architecture.} We use ResNet50~\cite{he2016deep} as the backbone feature encoder for all our experiments. Moreover, we initialize it on the source with weights pre-trained on ImageNet~\cite{deng2009imagenet}. For robust source training, we use weights obtained after adversarial training\footnote{https://github.com/MadryLab/robustness} on ImageNet. We maintain non-overlapping training, validation and test splits created randomly and evaluate the performance of all methods and tasks on the test split while using the validation split for model selection. For the VisDA-C dataset, we follow the established standard protocol~\cite{peng2017visda} by training our source models on synthetic images and adapting the models on the real images. All the experiments were conducted using the PyTorch framework~\cite{paszke2019pytorch}.
    
    \noindent\textbf{Implementation Details.} We keep the batch size fixed to 64 for all the datasets, tasks and methods. The learning rate is set to $10^{-3}$ for the classifier and the feature bottleneck layers while the backbone is trained at a slower rate of $10^{-5}$ using the Adam~\cite{kingma2014adam} optimizer. We use early stopping in all training-runs with a stop patience of 5. For generating adversarial examples we set the number of PGD~\cite{madry2017towards} steps to 20, attacking under the $l_{\infty}$ norm ($\epsilon = 4/255$) with a relative step size equal to $0.1/0.03$. Given the large size of Vis-DA, the source model reaches high-accuracy in just 2 epochs and the adaptation process is performed for 5 epochs. For all other datasets, we train the source model for 20 epochs and run adaptation for 10 epochs. The loss components weights $\alpha = 1.0$, $\beta = 0.3$, are borrowed from~\cite{liang2020we} and $\gamma = 0.2$.

    \noindent\textbf{Three Cases of  Our Method.} Recall that we also account for the case where only a single source model is available, as described in Section~\ref{subSec:singleModel}. When only the standard source model is available, we initialize the encoder with weights obtained after adversarial training on ImageNet, while the classifier is initialized randomly. This is done due to absence of a corresponding robust model in the source for initialization. To distinguish among the three scenarios, we refer to our method as Ours (robust source), Ours (standard source) or Ours (both) when only robust, only standard or both the models are available in the source domain.

\subsection{Baselines}
Since,to the best of our knowledge, there is no previous work on robust domain adaptation, we construct two baselines. The baselines use state-of-the-art domain adaptation approaches~\cite{tzeng2017adversarial, liang2020we} that we adapt to use adversarial training in the source domain.

The first adapted approach is Adversarial Discriminative Domain Adaptation (ADDA) by~\cite{tzeng2017adversarial} which, in addition to the data our approach requires, also uses source data in the adaptation phase. We perform adversarial training~\cite{madry2017towards} in the source domain and follow the target adaptation protocol as described in~\cite{tzeng2017adversarial}.

The second method we use for comparison is Source Hypothesis Transfer (SHOT)~\cite{liang2020we}. This approach is, similar to our approach, a source-free method, and thus, does not require access to source data during adaptation. We again modify this approach to use adversarial training~\cite{madry2017towards} in the source domain and subsequently follow the adaptation strategy as described in~\cite{liang2020we}.

Most of the source-free UDA methods require image/feature generation~\cite{li2020model,kurmi2021domain,kundu2020universal} which are difficult to scale while ensuring robustness on large datasets like VisDA-C~\cite{peng2017visda}. Other recently introduced approaches~\cite{sahoo2020unsupervised, sun2019test, wang2020fully} that are designed for pixel level corruptions~\cite{hendrycks2018benchmarking} do not extend well to more complex domain adaptation tasks we present in this paper.

\subsection{Results} We evaluate all our methods along with the introduced baselines on all four datasets, Office-31, Office-home, PACS and VisDA-C. We report the averages over all classes and adaptation tasks for all datasets. An exception is the VisDA-C dataset, where we follow the standard protocol and report the per-class average for Synthetic (S) to Real (R). The accuracies on adversarial attacks are visualized in Fig.~\ref{fig:results} which shows that all our methods outperform the baselines. More detailed results on all the datasets are presented in Table~\ref{table:results}.

\begin{figure*}[t!]
    \centering
    \includegraphics[scale=0.15]{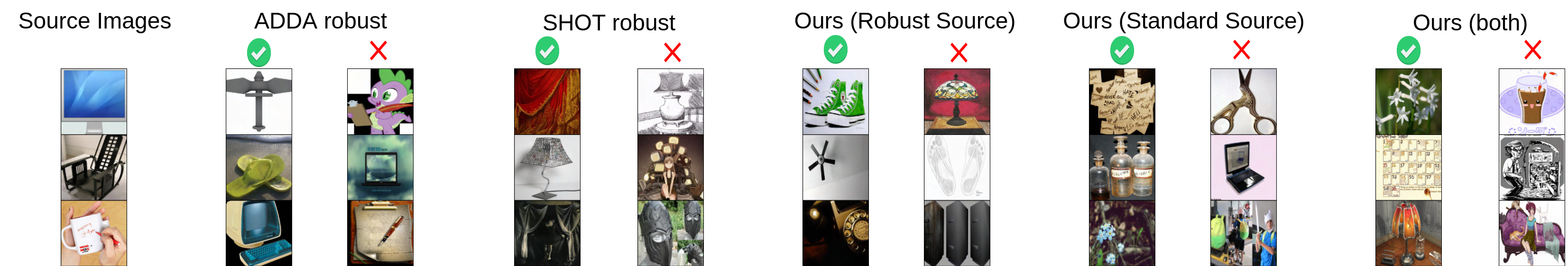}
    \caption{Sample adversarial images in the target (Art) domain of Office-home. The figure shows the correctly classified and misclassified images by the target model for each of the method. The source model was trained on Real-world (Rw) images.}
    \label{fig:samples}
\end{figure*}

All introduced methods perform consistently better than the baselines on adversarial images on all datasets. Besides having a good performance in the case of adversarial attacks, our models also perform competitively on clean samples.
On the PACS dataset, our (standard source) method outperforms all others, both in clean and adversarial accuracy. 
On Office-31 and Office-home, our method (both) improves robust accuracy by 15.9\% and 11.9\% respectively while only loosing 0.6\% and 0.7\% clean accuracy compared to the best non-robust model. Overall, our two best approaches (standard source and both) significantly improve adversarial accuracy while only reducing clean accuracy slightly (max -4.1\% on VisDA-C). Fig.~\ref{fig:samples} shows randomly selected adversarial images from the target domain (Art) which are classified correctly and incorrectly by the five different robust models adapted from the Real-world (Rw) domain in Office-home.

It is important to note that the clean accuracy for both ADDA and SHOT drops considerably if they are directly trained robustly. This is in line with the general observation that robust models tend to hurt the performance on clean samples~\cite{zhang2019theoretically}.

In two of the datasets (Office-31 and Office-home) our method which utilizes both standard and robust source models performs best. This is switched in the other two datasets, where our method which only requires the standard source model is better. To further analyze this behavior, we create two subsets of the data by only keeping all the images that belong to the first\footnote{In alphabetical order of the class labels, which is not related to the class complexity.} 10 and 32 classes respectively, both in the source and the target domain. This is done to ensure that the number of samples per class remain the same in all the three cases.

\begin{figure}
    \centering
    \includegraphics[scale=0.3]{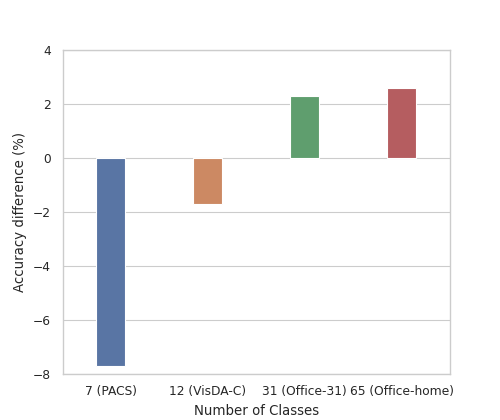}\hspace{2mm}
    \includegraphics[scale=0.3]{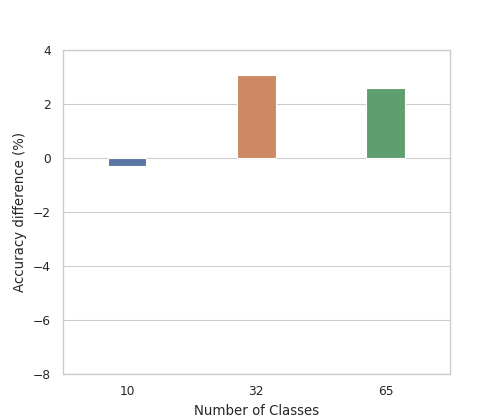}
    \caption{Performance of our method that use both source models relative to our method using only standard source model. The plot on the left shows the comparison on all the four dataset while on the right compares performance on Office-home by varying the number of classes.}
    \label{fig:behaviour}
\end{figure}

Results for these adaptation tasks are illustrated in Fig.~\ref{fig:behaviour} where we compare the method using only the standard model against the method using both source models. Fig.~\ref{fig:behaviour} indicates that in the case of few classes having only standard source model suffices for robust adaptation. However, if a larger set of classes needs to be handled, it is better to make use of both the standard and robust source model and follow the procedure as described in Section~\ref{fig:model_sep}.

\subsection{Ablation Study}
We study the impact of each of the components in our model on the PACS dataset in Table~\ref{table:ablation}. Removing the contrastive loss from both target model and the target robust model reduces the average performance. Similarly, the target accuracy decreases without the entropy minimization term or the diversity loss. The absence of cross-entropy loss calculated with help of pseudo-labels also makes it hard for the model to adapt well. Furthermore, we find that generating adversarial images using pseudo-labels also plays a significant role in improving the robust accuracy of the model.

\begin{figure*}[h]
     \centering
     \begin{subfigure}[b]{0.24\textwidth}
         \centering
         \includegraphics[width=\textwidth]{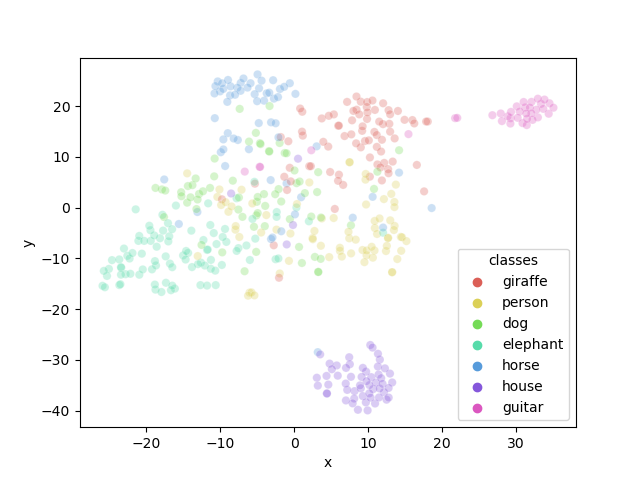}
         \caption{Before Adaptation}
         \label{fig:source}
     \end{subfigure}
     \begin{subfigure}[b]{0.24\textwidth}
         \centering
         \includegraphics[width=\textwidth]{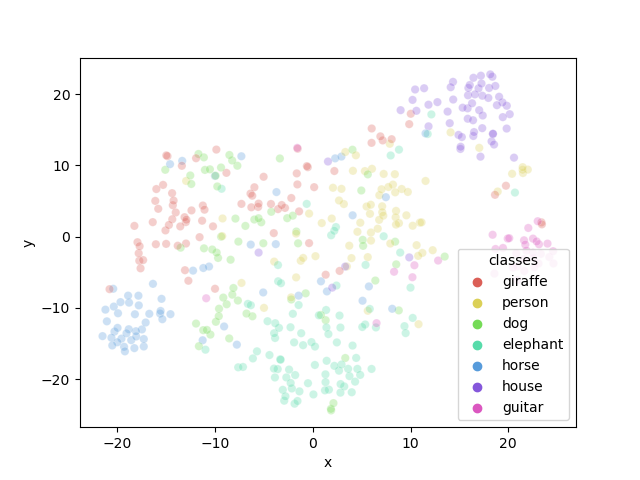}
         \caption{No pseudo-labels}
         \label{fig:no_labels}
     \end{subfigure}
     \hfill
     \begin{subfigure}[b]{0.24\textwidth}
         \centering
         \includegraphics[width=\textwidth]{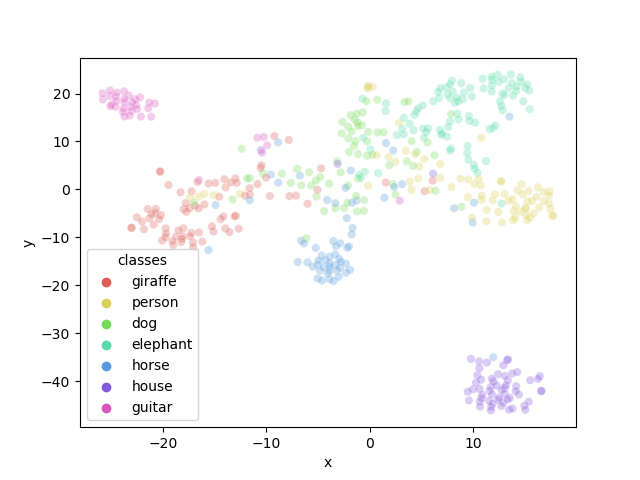}
         \caption{Ours (robust source)}
         \label{fig:robust_labels}
     \end{subfigure}
     \hfill
     \begin{subfigure}[b]{0.24\textwidth}
         \centering
         \includegraphics[width=\textwidth]{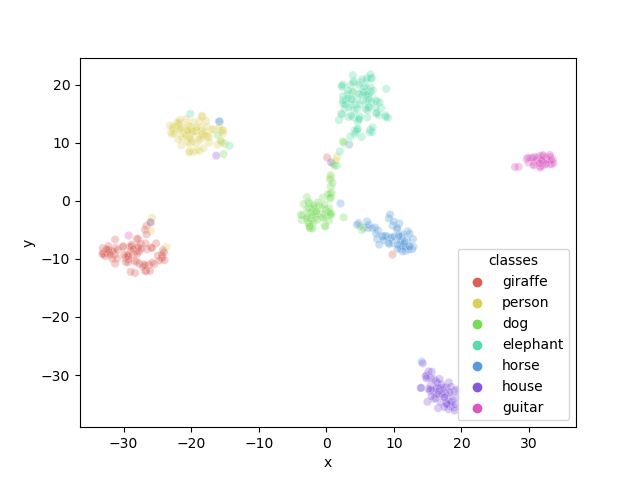}
         \caption{Ours (standard source)}
         \label{fig:non-robust_labels}
     \end{subfigure}
        \caption[Feature visualization for Art to Cartoon]{Impact of pseudo-labels on PACS with Art painting as the source and Cartoon as the target domain. The subplots clearly show that the features learned become more and more discriminative, forming better clusters as we introduce pseudo-labels and obtain them from the standard target model instead of the robust target model.}
        \label{fig:tsne-pseudo}
\end{figure*}

Recall that we require pseudo-labels to calculate the cross-entropy and contrastive loss and generate adversarial images in the target domain. To analyze the impact of pseudo-labels, we visualize the features of the adversarial images for the adaptation from Art (A) to Cartoon (C) on PACS under four different scenarios. We make use of PCA followed t-SNE~\cite{van2008visualizing} for dimensionality reduction.  Fig.~\ref{fig:source} shows the target features of the robust source model. Next, we perform domain adaptation without using any pseudo-labels and plot the encoder features as shown in Fig.~\ref{fig:no_labels}. In the next setting, we use pseudo-labels generated from a robust target model instead. Fig.~\ref{fig:robust_labels} shows that adversarial test images in this scenario form better clusters in the feature space. Finally, we compare it with our protocol, where we generate pseudo-labels from the standard target model to train the robust target encoder in Fig.~\ref{fig:non-robust_labels}. Fig.~\ref{fig:tsne-pseudo} clearly demonstrates that the learned features become more and more discriminative, forming better clusters as we introduce pseudo-labels and obtain them from the standard target model instead of the robust target model.

\section{Discussion}

Based on our experimental evaluations, we attempt to answer some key questions. We believe that our answers to these questions help to better understand the outcome of our study as well as the problem addressed in this paper.   

\noindent\textbf{Do robust models transfer robustly?}--Yes. Beside the proposed method, our baselines also allow us conclude that the robust models indeed transfer robustly. In particular, two baselines, SHOT robust and ADDA robust do not even use the adversarial examples in the target domain.
The performance of these methods on the adversarial examples are noteworthy.  This observation is in accordance to the existing works~\cite{shafahi2019adversarially, utrera2020adversarially, salman2020adversarially}, although in different settings. In the setting of this paper, the robust source models are found to be very useful for datasets with many classes.

\noindent\textbf{Which pseudo-labels to use?}--Non-robust. Our experiments demonstrate the clear benefit of using non-robust pseudo labels  for robustness in the target domain. Please, refer to Sec.~\ref{subSec:robustTraning} and \ref{subSec:singleModel} for more details. It goes without saying, non-robust pseudo-labels are preferred when non-robust source models are available. 

\noindent\textbf{Which model to transfer?}--Robust.
Provided a good transfer of non-robust models to the target, it has been observed that the robustness can be achieved by generating the adversarial examples in the target. Such robustness however fully relies on the pseudo-labels alone. 
We observed that for datasets with few class such transfer is often is not a problem. 
However, as the number of classes increases, the transfer of non-robust models followed by the robust training is not a good idea. Please, refer to Fig.~\ref{fig:behaviour} for robust and non-robust models transfer for increasing number of classes. Such behaviour can be attributed to the following: as the number of classes increases, the chanced of pseudo-labels being incorrect in the target becomes higher. As the transfer of robust source model does not fully rely only on the pseudo-labels, we suggest to adapt the roust source model. This suggestion is however, meant to be followed for the two models case. Otherwise, we recommend to transfer the non-robust source model followed by robust target training (using the method proposed in this paper).    

\noindent\textbf{What makes any given model better?}--Contrastive loss. The use of contrastive loss for the addressed problem is found to be very helpful in all three cases of the model availability presented in Sec.~\ref{subSec:singleModel}. This can be observed in Tab.~\ref{table:ablation} and ~\ref{table:results}. Note that the baseline SHOT robust differs from our method with robust source in terms of the contrastive feature learning. Please, refer Sec.~\ref{subSec:nonrobustTraning} for the details.   

\noindent\textbf{How do I design the transfer protocol?}--Transfer two models. When the availability of source models is not a problem, we suggest to use two models as presented in Fig.~\ref{fig:model-abs-se} and Algo.~\ref{alg:ourAlgorithm}. This may be particularly important, when designing the model transfer protocol is possible.

\noindent\textbf{Do I need to keep two models after transfer?}--No. Only using the transferred robust model will offer the adversarial and clean accuracy  of Tab.~\ref{table:results}. The non-robust model is only used to generate more reliable pseudo-labels, for adversarial examples during robust training in the target domain.   

\section{Conclusion}
We study three different cases of model availability for the unsupervised robust domain adaptation without source data. These cases were chosen to model practical scenarios. In all the three cases, we obtained very promising results, thanks to the proposed method. Our extensive study shows that the transfer of both robust and standard model is often the best choice for the robustness in the target domain. Overall, the non-robust pseudo-labels and contrastive feature learning strategies are found to be very effective, when combined with the existing model transfer methods. In future, we will explore single source models that perform both robust and non-robust predictions, in a multi-tasking fashion. This will avoid sharing two models trained on the the source data.

{\small
\bibliographystyle{ieee_fullname}
\bibliography{egbib}
}

\end{document}